# Multi-View Fuzzy Clustering with The Alternative Learning between Shared Hidden Space and Partition

Zhaohong Deng, *Senior Member*, *IEEE*, Chen Cui, Peng Xu, Ling Liang, Haoran Chen, Te Zhang, Shitong Wang

*Abstract*—As the multi-view data grows in the real world, multi-view clustering has become a prominent technique in data mining, pattern recognition, and machine learning. How to exploit the relationship between different views effectively using the characteristic of multi-view data has become a crucial challenge. Aiming at this, a hidden space sharing multi-view fuzzy clustering (HSS-MVFC) method is proposed in the present study. This method is based on the classical fuzzy c-means clustering model, and obtains associated information between different views by introducing shared hidden space. Especially, the shared hidden space and the fuzzy partition can be learned alternatively and contribute to each other. Meanwhile, the proposed method uses maximum entropy strategy to control the weights of different views while learning the shared hidden space. The experimental result shows that the proposed multi-view clustering method has better performance than many related clustering methods.

*Index Terms*—multi-view data; clustering; maximum entropy; shared hidden space

## I. Introduction

Multi-view data has become more and more common in real world. An important characteristic of multi-view data is that different views often provide compatible and complementary information. For example, a dataset of bank customer can be categorized into views of population information, account information and consumption information. In different application fields, many multi-view algorithms [1-6] have been proposed to make full use of multi-view data, such as multi-view cooperation learning in clustering. In recent years, besides the above theoretical development, multi-view learning has many achievements in practical application, especially in classification and clustering. In the present study, we focus on the challenge that exists in multi-view clustering and propose a potential solution that enhances the clustering performance.

Different from classical single view clustering method, multi-view clustering method usually integrates information from different views and is expected to get better clustering performance. Many multi-view clustering methods have been proposed in relevant literature [1-6] and most of them are classified into two categories:

i. Multi-view clustering methods based on single view clustering methods, such as K-means [7, 8], Fuzzy C-means (FCM) [9,10], Maximum Entropy Clustering (MEC) [11, 12] and Possibilistic C-means (PCM) [13, 14]. Such approach usually considers each view independently and treats each of them as an independent clustering task, followed by ensemble methods [15, 16] to achieve the final clustering result. However, such strategy is likely to cause poor results or unsteadiness of the algorithm because of potential high deviation of a certain view or significant discrepancies between the results of each view.

ii. Multi-view clustering methods based on algorithms with multi-view learning mechanics, such as algorithms designed based on space transformation methods [1, 5, 17, 18]. Recently, Liu et al. [19] presented a novel tensor-based framework for integrating heterogeneous multi-view data in the context of spectral clustering. Zhang et al. [20] proposed low-rank tensor constrained multi-view subspace clustering which regards the subspace representation matrices of different views as a tensor equipped with a low-rank constraint. The multi-linear relationship among multi-view data is taken into account through their tensor-based strategy. In order to deal with large-scale data clustering problems, a new robust large-scale multi-view clustering method [21] was proposed to integrate multiple representations of large scale data. Li et al. [22] presented partial multi-view clustering in the case that every view suffers from the missing of some data and results in many partial examples. Wang et al. [23] proposed a multi-view learning model to integrate all features and learn the weight for every feature with respect to each cluster individually via new joint structured sparsity-inducing norms. Some researchers have proposed multi-view clustering ensemble learning that combines different ensemble techniques for multi-view clustering [24-26].

Among the existing techniques used for constructing the shared subspace for multiple views, non-negative matrix factorization has attracted extensive attention. [36] first proposed to learn a shared hidden space for multiple views using NMF. The method proposed in [37] mainly aimed at tackling the problems in the scene of incomplete multiple views, the weighted NMF is constructed with the $L_{1,2}$ regularizations for the multi-view clustering. [38] introduced multi-manifold regularization into the NMF-based multi-view clustering. For the application of object recognition, the graph-regularization was also introduced into NMF-based multi-view clustering [39]. Similar to the method proposed in [38], the correlation constraint was considered as well in [40]. Although these methods have developed NMF-based multi-view clustering from different aspects, a common disadvantage of them is that the pro-

This work was supported in part by the Outstanding Youth Fund of Jiangsu Province (BK20140001), by the National Key Research Program of China under grant 2016YFB0800803, the National Natural Science Foundation of China (61272210). (Corresponding author: Zhaohong Deng)

Zhaohong Deng, Chen Cui, Peng Xu, Ling Liang, Te Zhang, Shitong Wang are with the School of Digital Media, Jiangnan University and Jiangsu Key Laboratory of Digital Design and Software Technology, Wuxi 214122, China (e-mail: dengzhaohong@jiangnan.edu.cn; 569398535@qq.com; 6171610015@stu.jiangnan.edu.cn; 6181611019@stu.jiangnan.edu.cn; ztcsrookie@163.com; wxwangst@aliyun.com)

H. Chen is with the Mathematics Department of the University of Michigan, Ann Arbor, MI 48109, USA (e-mail: chenhr@umich.edu)

cedure of constructing the hidden space is uncoupled with the procedure of calculating the clustering centers.

Different from existing multi-view clustering methods, a new shared hidden space learning method is proposed in the present study to exploit the relationship between different views along with the optimization of the clustering centers. This method considers information from different views while expecting to find essential attributes of clustering objects for more reasonable results. In detail, Non-negative Matrix Factorization (NMF) is used to learn the shared hidden space. Based on NMF, the attributes matrix is decomposed into the product of basis matrix of single view and coefficient matrix shared by different views. Besides, in order to adaptively fuse the information for each view, maximum entropy is introduced to adjust the weight of each view. Applying the above strategy into the classical Fuzzy C-means framework, the algorithm hidden-space-sharing multi-view fuzzy clustering (HSS-MVFC) is proposed.

The main contributions of this paper can be highlighted as follows:

1) An approach to extract shared hidden information among the visible views using non-negative matrix factorization.

2) A hidden space sharing multi-view fuzzy clustering method based on FCM.

3) Introduction of maximum entropy to adjust the weight of each view.

4) Validation of the proposed HSS-MVFC using extensive experiments

The rest of this paper is organized as follows. Section II briefly reviews the concepts and principles of classical FCM and NMF. Section III proposes a strategy of hidden space sharing multi-view fuzzy clustering. Section IV presents the experimental results. Finally, Section V draws conclusions and points out some future research directions.

## II. BACKGROUND KNOWLEDGE

Our study is closely related to both FCM and NMF. The framework of FCM is the basic framework of the proposed method and that of NMF is used to construct the shared hidden view. Besides, some established multi-view clustering methods based on FCM is also briefly reviewed.

### A. Single View FCM

FCM is a classical single view clustering method. Given a dataset $\{\mathbf{x}_1, \mathbf{x}_2, ..., \mathbf{x}_n\}$, $\mathbf{x}_i = (x_1, x_2, ..., x_d)^T \in \mathbf{R}^d$ (where $n$ is the number of samples, $1 \leq i \leq n$, $d$ is the number of attributes), the objective of FCM is to find a sample fuzzy partition matrix $\mathbf{U} = [u_{ij}]_{c \times n}$ (where $c$ is the number of clusters with $1 \leq i \leq c$ and $1 \leq j \leq n$) and the central matrix of clusters $\mathbf{V} = [\mathbf{v}_1, \cdots, \mathbf{v}_c]$, $\mathbf{v}_i = (v_{i1}, \cdots, v_{id})^T$. The objective function of FCM is

$$\begin{cases} J_{FCM}(\mathbf{U}, \mathbf{V}) = \sum_{i=1}^{c} \sum_{j=1}^{n} u_{ij}^m \| \mathbf{x}_j - \mathbf{v}_i \|^2 \\ s.t. \ u_{ij} \in [0,1], \sum_{i=1}^{c} u_{ij} = 1, \ 1 \leq j \leq n \end{cases} \quad (1)$$

where $\mathbf{v}_i$ is the center of cluster $i$, $\mathbf{x}_j$ represents sample $j$, $u_{ij}$ is the membership of sample $j$ in cluster $i$, $m$ is the fuzzy index and $m > 1$. By Lagrange optimization, we can obtain the update rules of $\mathbf{v}_i$ and $u_{ij}$ as follow:

$$\mathbf{v}_i = \sum_{j=1}^{n} u_{ij}^m \mathbf{x}_j \bigg/ \sum_{j=1}^{n} u_{ij}^m \quad (2)$$

$$u_{ij} = 1 \bigg/ \sum_{k=1}^{c} \left[ \frac{\| \mathbf{x}_j - \mathbf{v}_i \|^2}{\| \mathbf{x}_j - \mathbf{v}_k \|^2} \right]^{\frac{2}{m-1}} \quad (3)$$

Using the above update rules, we can obtain the final partition matrix and clustering centers.

Finally, the fuzzy partition matrix of single-view data can be obtained by the above optimization strategy. When facing multi-view clustering, without considering the relevance between different views, the most direct way is to use (1) to get partition matrix $\mathbf{U}_k$ (where $k$ denotes the $k$th view, $1 \leq k \leq K$) of each view independently. Then, ensemble learning technique is introduced to integrate $\mathbf{U}_k$ into $\tilde{\mathbf{U}}$ which is a space partition matrix with global description capability.

The above method provides a feasible way for single-view clustering when facing multi-view scene. However, it ignores the relationship among the multiple views, and hence might not get the best clustering performance.

### B. Co-FKM

As we may know, FCM [9, 10] is a classical fuzzy clustering methods. Just like many existing multi-view clustering methods such as Co-FKM [27], Co-FCM [28] and WV-CO-FCM [6], the proposed HSS-MVFC method will continue to treat FCM as the basic framework. Among those methods, Co-FKM is a classical multi-view fuzzy clustering technique. A brief review of Co-FKM is given as follows.

Given dataset $\mathbf{X} = \{\mathbf{x}_1, \mathbf{x}_2, ..., \mathbf{x}_n\}$, $\mathbf{x}_{j,k}$ denotes the kth view data of sample $\mathbf{x}_j$. $\mu_{ij,k} \in [0,1]$ is the membership of $\mathbf{x}_{j,k}$ in cluster $i$, $\mathbf{v}_{i,k}$ denotes the center of cluster $i$, $d_{ij,k}$ denotes the Euclidean distance of sample $j$ to center $\mathbf{v}_i$ in view $k$. The objective function of Co-FKM is as follows:

$$J_{\text{Co-FKM}}(\mathbf{U}, \mathbf{V}) = \sum_{k=1}^{K}(J_{FKM}(\mathbf{U}_k, \mathbf{V}_k)) + \eta \Delta \quad (4)$$

$$s.t. \ \mu_{ij,k} \in [0,1] \text{ and } \sum_{i=1}^{C} \mu_{ij,k} = 1 \ 1 \leq j \leq n \ 1 \leq k \leq K$$

where,

$$J_{FKM}(\mathbf{U}_k, \mathbf{V}_k) = \sum_{i=1}^{C} \sum_{j=1}^{N} [\mu_{ij,k}^m \| \mathbf{x}_{j,k} - \mathbf{v}_{i,k} \|^2] \quad (5)$$

$$\Delta = \frac{1}{K-1} \sum_{k'=1, k' \neq k}^{K} \sum_{i=1}^{C} \sum_{j=1}^{N} (\mu_{ij,k'}^m - \mu_{ij,k}^m) \| \mathbf{x}_{j,k} - \mathbf{v}_{i,k} \|^2 \quad (6)$$

$K$ denotes the number of views and $\eta$ is a cooperative learning parameter which controls the membership division of each view.

The first term of (4), i.e., $\sum_{k=1}^{K}(J_{FKM}(\mathbf{U}_k, \mathbf{V}_k))$ is an empirical loss term. This term minimizes $J_{FKM}(\mathbf{U}_k, \mathbf{V}_k)$ by each view, and can be regarded as clustering by each view independently. The second term $\Delta$ in (6) is a penalty term. This term tries to make the outputs from different views to be consistent while the algorithm is converging, which may help to realize the generalization of each view.

Substitute (6) into (4) and we get the following objective function:

$$J_{\text{Co-FKM}}(U,V) = \sum_{k=1}^{K}\sum_{i=1}^{C}\sum_{j=1}^{N}\left[\tilde{\mu}_{ij,k,\eta}\left\|\mathbf{x}_{j,k}-\mathbf{v}_{i,k}\right\|^{2}\right] \quad (7)$$

Where

$$\tilde{\mu}_{ij,k,\eta} = (1-\eta)\mu_{ij,k}^{m} + (\eta/(K-1))\sum_{k'=1,k'\neq k}^{K}\mu_{ij,k'}^{m}$$

$\eta$ is a cooperative learning parameter that adjusts the membership of each view. $\tilde{\mu}_{ij,k,\eta}$ is the weighted mean of current view's membership and the rest views' membership.

According to the optimization strategy of classical FCM, the update formula of $\mu_{ij,k}$ and $\mathbf{v}_{i,k}$ are as follows:

$$\mathbf{v}_{i,k} = \sum_{j=1}^{N}\tilde{\mu}_{ij,k,\eta}\mathbf{x}_{j,k} \Big/ \sum_{j=1}^{N}\tilde{\mu}_{ij,k,\eta} \quad i=1,2,...,C \quad (8)$$

$$\mu_{ij,k} = 1\Big/\sum_{h=1}^{C}\left[\frac{(1-\eta)d_{ij,k}^{2}+\dfrac{\eta}{K-1}\sum_{k'=1,k'\neq k}^{K}d_{ij,k'}^{2}}{(1-\eta)d_{hj,k}^{2}+\dfrac{\eta}{K-1}\sum_{k'=1,k'\neq k}^{K}d_{hj,k'}^{2}}\right]^{\frac{1}{m-1}} \quad (9)$$

$$i=1,2,...,C, j=1,2,...,N$$

where $d_{ij,k}$ is the distance of view $k$. The specific form of $d_{ij,k}$ is $d_{ij,k} = dist(\mathbf{x}_{j,k}-\mathbf{v}_{i,k}) = \left\|\mathbf{x}_{j,k}-\mathbf{v}_{i,k}\right\|$.

The final fuzzy membership partition matrix of each view can be obtained using the above update strategy. [18] [29] use geometric mean of fuzzy membership obtained by each view to reflect the overall division results:

$$\hat{\mu}_{ij} = \sqrt[k]{\prod_{k\in K}\mu_{ij,k}} \quad (10)$$

Spatial partition results of the whole data can be obtained by the defuzzification of the results of (10). This spatial partition results provide a powerful space division reference for making decisions about the data.

The above algorithm realizes the cooperation between each view and has better performance when facing multi-view data compared with single view clustering algorithms. However, the above algorithm still has some remaining challenges. For example, the cooperation learning principle is too simple and has weak physical interpretation.

### C. NMF

Non-negative matrix factorization (NMF) is a dimension reduction technique. It has a wide application in the field of pattern recognition, image engineering and other research fields. By using non-negative matrix factorization to extract the features of data, the dimension of non-negative data is reduced. Given a non-negative data matrix $\mathbf{X} = [\mathbf{x}_1,\mathbf{x}_2,...,\mathbf{x}_N]$ with $m$ dimensional features and $N$ samples, NMF aims to find two non-negative matrix factors $\mathbf{P}\in R_{+}^{m\times r}$ and $\mathbf{H}\in R_{+}^{r\times N}$ where $r$ denotes the dimension after the desired dimension reduction, $\mathbf{P}$ and $\mathbf{H}$ denote the basis matrix and coefficient matrix respectively, so that $\mathbf{X}\approx \mathbf{PH}$, that is, their inner product can approximatively represent $\mathbf{X}$. Therefore, NMF can be attributed to optimizing the following objective function problem, i.e.

$$\min_{\mathbf{W},\mathbf{H}}\left\|\mathbf{X}-\mathbf{PH}\right\|_{F}^{2} \quad (11)$$
$$s.t. \ \mathbf{P}\geq \mathbf{0},\mathbf{H}\geq \mathbf{0}$$

where $\left\|\cdot\right\|_{F}$ represents the Frobenius norm.

## III. HIDDEN SPACE SHARING MULTI-VIEW FUZZY CLUSTERING

As described earlier, a key problem for multi-view clustering is how to exploit the compatible information that exists between each view. A reasonable hypothesis is that there exists a hidden view shared by multiple views that reflects such information. In this paper, we use NMF to get the shared hidden view and propose Hidden Space Sharing Multi-View-Fuzzy Clustering(HSS-MVFC). The flowchart of the HSS-MVFC algorithm is shown in Fig.1.

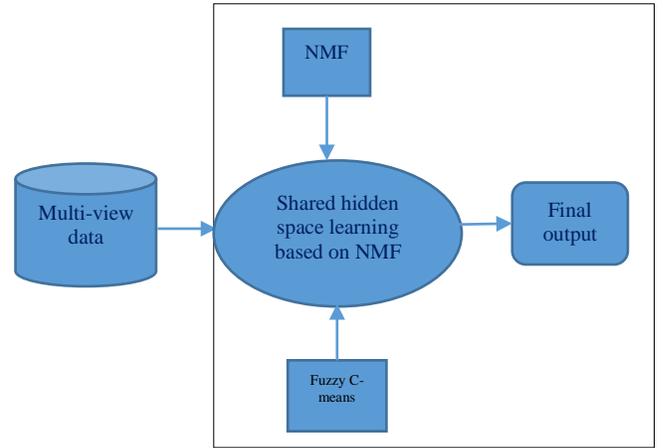

Fig.1. The flowchart of the HSS-MVFC algorithm

### A. Shared Hidden Space Learning based on NMF

To overcome the challenge that single view algorithms are unable to make full use of compatible information that exists in different views, NMF is used. Based on NMF, each view data can be factorized into a base matrix P and coefficient matrix. Specifically, this process can be represented as the following optimization problem:

$$\sum_{k=1}^{K}\|\mathbf{X}^{k}-\mathbf{P}^{k}\mathbf{H}^{k}\|_{F}^{2} \quad (12)$$
$$s.t. \ \mathbf{P}^{k},\mathbf{H}^{k}\geq \mathbf{0}$$

where $K$ is the number of views, $\mathbf{X}^k =[\mathbf{x}_1^k, \mathbf{x}_2^k,...,\mathbf{x}_n^k]$ represents the $k$th view data, $n$ is the number of samples, $\mathbf{P}^k =[\mathbf{p}_1^k, \mathbf{p}_2^k,\cdots,\mathbf{p}_r^k] \in \mathbf{R}^{m_k \times r}$ represents the base matrix of view $k$, $m^k$ is the number of attributes of view $k$ in original space, $r$ is the dimensions of view $k$ data in the low-dimensional space ( $1 \leq r \leq \min\{m_1, m_2, ..., m_K, n\}$ ). $\mathbf{H}^k =[\mathbf{h}_1^k, \mathbf{h}_2^k,...,\mathbf{h}_n^k] \in \mathbf{R}^{r \times n}$ is the coefficient matrix of view $k$ and $\|.\|_F$ represents Frobenius norm.

For a certain set of multi-view data, although the characteristic spaces of any two views are different, they describe the same instance. Thus, the data from view $k$ and view $j$ can be regarded as maps from original attributes to view $k$ and view $j$. The key problem then is how to get the shared hidden space data $\mathbf{h}_i$.

As shown in (11), if $\mathbf{H} = \mathbf{H}^1 = \mathbf{H}^2 = \cdots = \mathbf{H}^K$, then $\mathbf{h}_i = \mathbf{h}_i^1 = \mathbf{h}_i^2 = ... = \mathbf{h}_i^K$ can be regarded as the shared hidden feature of sample $i$.

$$\sum_{k=1}^{K} \| \mathbf{X}^k - \mathbf{P}^k \mathbf{H} \|_F^2 \quad (13)$$
$$s.t. \quad \mathbf{P}^k, \mathbf{H} \geq 0$$

### B. Multi-view Adaptive Weighting based on Shannon Entropy

Most established multi-view clustering algorithms usually assume each views' contributions to the final results are equal. However, in reality the weight of different views in a same clustering task varies. A more reasonable approach is to assign different weights to different views according to their separable character, but it's hard to do so manually. To address this problem, we introduce maximization of Shannon entropy strategy [11] to realize the adaptive weighting.

Let $\sum_{k=1}^{K} w_k = 1$ and $w_k \geq 0$, where $w_k$ represents the weight of view $k$, we can get the weight vector $\mathbf{w} = (W_1, \cdots, W_K)^T$. The weights can be regarded as probability distribution and the uncertainty of the distribution can be represented by Shannon entropy:

$$f(w_k) = -\sum_{k=1}^{K} w_k \ln w_k \quad (14)$$

The Shannon entropy term is used to determine the optimal weights for different views. Maximizing $-\sum_{i=1}^{K} w_k \ln w_k$ will make $w_k$ close to each other [30], reducing the risk that a certain view dominates the final output.

### C. Object Function

The Lagrangian method is used for solving $p_g^k$ and $w_k$ in (14), and the updating rules are:

Using the strategies described in section III.A and section III.B, based on the framework of FCM, the objective function of the proposed HSS-MVFC is as follows:

$$J(\mathbf{U},\mathbf{V},\mathbf{P}^k,\mathbf{H},\mathbf{w}) = \min_{\mathbf{U},\mathbf{V},\{\mathbf{P}^k\}_{k=1}^{K},\mathbf{H},\mathbf{w}} \sum_{l=1}^{c} \sum_{i=1}^{n} u_{li}^m \| \mathbf{h}_i - \mathbf{v}_l \|^2$$
$$+ \lambda \sum_{k=1}^{K} w_k \| \mathbf{X}^k - \mathbf{P}^k \mathbf{H} \|_F^2 + \eta \sum_{k=1}^{K} w_k \ln w_k \quad (15)$$
$$s.t. \begin{cases} \sum_{l=1}^{C} u_{li} = 1, \ 0 \leq u_{li} \leq 1 \\ \sum_{k=1}^{K} w_k = 1, \ \leq 0 w_k \leq 1 \\ \{\mathbf{P}^k\}_{k=1}^{K}, \mathbf{H} \geq 0 \end{cases}$$

Here,
$\mathbf{U} = [u_{ij}]_{c \times n}$ is partition matrix, $1 \leq i \leq c, 1 \leq j \leq n$.
$\mathbf{X}^k = [\mathbf{x}_1^k, \cdots, \mathbf{x}_n^k]$ is the kth view data, $1 \leq i \leq K$
$\mathbf{V} = [\mathbf{v}_1, \cdots, \mathbf{v}_c]$ contains centers of the shared hidden space.
$\mathbf{P}^k \in \mathbf{R}^{m_k \times r}$ is the base matrix of view $k$, where $r$ represents the dimension of the shared low-dimensional space.
$\mathbf{H} = [\mathbf{h}_1, \mathbf{h}_2,...,\mathbf{h}_n] \in \mathbf{R}^{r \times n}$ is the coefficient matrix shared by $K$ views.
$\mathbf{W} = (w_1, w_2, ..., w_k)^T$ is the weight vector of $K$ views.
$\lambda, \eta$ are regularization parameters.

There are three terms in (15). The first term is used to realize fuzzy cluster based on shared hidden space. The second term is used to learn the shared hidden space, and the third term is used to realize adaptive control of the importance of each view during the hidden space learning process.

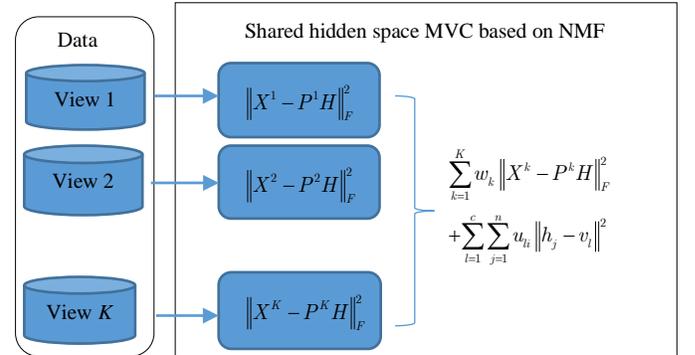

Fig 2 The Framework of HSS-MVFC

### D. Optimization Solution

(15) is a nonconvex optimization problem that can be solved by iterative optimization strategy. The optimization of (15) can be translated into solving the following five minimizer problems:

1. Problem $P_1$: Fix $\mathbf{V} = \hat{\mathbf{V}}$, $\mathbf{P}^k = \hat{\mathbf{P}}$, $\mathbf{H} = \hat{\mathbf{H}}$, $\mathbf{w} = \hat{\mathbf{w}}$ and solve $P_1(\mathbf{U}) = J(\mathbf{U},\hat{\mathbf{V}},\hat{\mathbf{P}},\hat{\mathbf{H}},\hat{\mathbf{w}})$

2. Problem $P_2$: Fix $\mathbf{U} = \hat{\mathbf{U}}$, $\mathbf{P}^k = \hat{\mathbf{P}}^k$, $\mathbf{H} = \hat{\mathbf{H}}$, $\mathbf{w} = \hat{\mathbf{w}}$ and solve $P_2(\mathbf{V}) = P(\hat{\mathbf{U}},\mathbf{V},\hat{\mathbf{P}}^k,\hat{\mathbf{H}},\hat{\mathbf{w}})$

3. Problem $P_3$: Fix $\mathbf{U} = \hat{\mathbf{U}}$, $\mathbf{V} = \hat{\mathbf{V}}$, $\mathbf{H} = \hat{\mathbf{H}}$, $\mathbf{w} = \hat{\mathbf{w}}$ and solve $P_3(\mathbf{P}^k) = P(\hat{\mathbf{U}},\hat{\mathbf{V}},\mathbf{P}^k,\hat{\mathbf{H}},\hat{\mathbf{w}})$

4. Problem $P_4$: Fix $\mathbf{U} = \hat{\mathbf{U}}$, $\mathbf{V} = \hat{\mathbf{V}}$, $\mathbf{P}^k = \hat{\mathbf{P}}^k$, $\mathbf{w} = \hat{\mathbf{w}}$ and solve $P_4(\mathbf{H}) = P(\hat{\mathbf{U}},\hat{\mathbf{V}},\hat{\mathbf{P}}^k,\mathbf{H},\hat{\mathbf{W}})$

5. Problem $P_5$: Fix $\mathbf{U} = \hat{\mathbf{U}}$, $\mathbf{V} = \hat{\mathbf{V}}$, $\mathbf{P}^k = \hat{\mathbf{P}}^k$, $\mathbf{H} = \hat{\mathbf{H}}$ and solve $P_5(\mathbf{w}) = P(\hat{\mathbf{U}}, \hat{\mathbf{V}}, \hat{\mathbf{W}}, \hat{\mathbf{H}}, \mathbf{w})$

The solving process of each sub-optimization problem can be described as follows:

1) sub-optimization problem $P_1$ and $P_2$ can be regarded as solving classical FCM, and the solution of $P_1$ and $P_2$ are as follows:

$$\mathbf{v}_l = \sum_{i=1}^{n} u_{li}^m \mathbf{h}_i \Big/ \sum_{i=1}^{n} u_{li}^m \tag{16}$$

$$u_{li} = 1 \Big/ \sum_{l'=1}^{c} (\frac{\|\mathbf{h}_i - \mathbf{v}_l\|^2}{\|\mathbf{h}_i - \mathbf{v}_{l'}\|^2})^{\frac{2}{m-1}}, \quad 1 \le i \le n \tag{17}$$

2) Sup-optimization problem $P_3$ is a nonnegative quadratic problem and is similar to classical NMF model. Based on the strategy used in [31], the following update rule can be used:

$$(\mathbf{P}^k)_{i,j}^{(t+1)} \leftarrow \left[ \frac{(\mathbf{X}^k \mathbf{H}^T)_{i,j}}{((\mathbf{P}^k)^{(t)} \mathbf{H} \mathbf{H}^T)_{i,j}} \right] (\mathbf{P}^k)_{i,j}^{(t)} \tag{18}$$

3) Sup-optimization problem $P_4$ can use gradient descent to optimize $\mathbf{H}$. The approach is to initialize $\mathbf{H}^{(0)}$ and let $(\mathbf{H})_{i,j}^{(t)}$ represent the element of $i$th row $j$th column, where $t$ is the number of iteration update times. The following formula is used to update $\mathbf{H}$:

$$(\mathbf{H})_{i,j}^{(t+1)} = \frac{\sum_{l=1}^{C} u_{li}^m (\mathbf{V})_{lj} + \lambda \sum_{k=1}^{K} w_k (\mathbf{P}^k \mathbf{X}^{(t)})_{i,j}}{\sum_{l=1}^{C} u_{li}^m (\mathbf{H})_{i,j}^{(t)} + \lambda \sum_{k=1}^{K} w_k (\mathbf{P}^k (\mathbf{P}^k)^T \mathbf{H}^{(t)})_{i,j}} (\mathbf{H})_{i,j}^{(t)} \tag{19}$$

$$1 \le j \le r, \ 1 \le i \le n$$

4) Set $D_k = \|\mathbf{X}^k - \mathbf{P}^k \mathbf{H}\|_F^2$ Sup-optimization problem $P_5$ can be optimized by Lagrangian multiplier strategy and we obtain the solution of $w_k$, the Lagrangian function is as follows:

$$L = \min_{\mathbf{U},\mathbf{V},\{\mathbf{P}^k\}_{k=1}^{K},\mathbf{H},\mathbf{w}} \sum_{l=1}^{c} \sum_{i=1}^{n} u_{li}^m \|\mathbf{h}_i - \mathbf{v}_l\|^2 + \lambda \sum_{k=1}^{K} w_k D_k + $$
$$\eta \sum_{k=1}^{K} w_k \ln w_k + \delta(\sum_{l=1}^{C} u_{li} - 1) + \gamma(\sum_{k=1}^{K} w_k - 1) \tag{20a}$$

By setting the partial derivative of (20a) with respect to $w_k$ to 0, $w_k$ can be expressed as:

$$w_k = \exp\left(\frac{-\lambda D_k - \eta - \gamma}{\eta}\right) = \exp\left(\frac{-\lambda D_k}{\eta}\right) \exp\left(\frac{-\eta - \gamma}{\eta}\right), \tag{20b}$$
$$1 \le k \le K$$

Substitute (20b) into the constraint condition in (15), the following equation can be obtained.

$$\exp\left(\frac{-\eta - \gamma}{\eta}\right) = \frac{1}{\sum_{k=1}^{K} \exp\left(\frac{-\lambda D_k}{\eta}\right)} \tag{20c}$$

The analytical solution of $w_k$ can be obtained by substituting (20c) into (20b), that is,

$$w_k = \frac{\exp\left(\frac{-\lambda D_k}{\eta}\right)}{\sum_{k'=1}^{K} \exp\left(\frac{-\lambda D_{k'}}{\eta}\right)}, \quad 1 \le k \le K \tag{20d}$$

Obviously, parameter $\eta$ will influence $w_k$, the weight of each view. If $\eta \to \infty$, all weights tend to be equal, i.e., each view is treated equally. If $\eta \to 0$, the most important view will play a decisive role and other views are most likely to be ignored. Thus the importance degree of each view can be controlled by adjusting $\eta$.

*E. Algorithm Description*

The pseudocode of the proposed HSS-MVFC can be described as follows.

TABLE I
FLOW CHART OF ALGORITHM HSS-MVFC

**Algorithm**: **HSS-MVFC**

**Input**: Multi-view dataset $\{\mathbf{X}^k\}, k = 1, \cdots, K$,
  clusters number $c$,
  regularization parameters $\lambda, \eta$
  maximum number of iterations $t_{\max}$.

**Output**: $\mathbf{U}, \mathbf{V}, \mathbf{P}^k, \mathbf{H}, \mathbf{w}$, where $k = 1,...K$.

**Procedure HSS-MVFC**:

1: Generate Nonnegative matrix $\mathbf{P}^k$ and $\mathbf{H}$, and clustering center matrix of hidden view $\mathbf{V}^{(0)}$, $w_k^{(0)} \leftarrow 1/K$ and $t \leftarrow 0$.

2: **Repeat:**

3:   Update $\mathbf{U}^{(t+1)}$ by (17);

4:   Update $\mathbf{V}^{(t+1)}$ by (16);

5:   Update $(\mathbf{P}^k)^{(t+1)}$ by (18);

6:   Update $\mathbf{H}^{(t+1)}$ by (19);

7:   Update $\mathbf{w}^{(t+1)}$ by (20d);

8:   $t \leftarrow t + 1$;

9:   Until (7) reaches a minimum or the number of iterations reaches $t_{\max}$.

10: **end repeat.**

*F. Computational Complexity*

The computational complexity of Algorithm HSS-MVFC in Table I is analyzed using big $O$ notation. Based on the above discussion, denote the number of views as $K$, the number of samples as $n$, the dimension of the shared hidden space as $r$, the number of clusters as $c$ and the maximum dimension of the original data is $m$. For the problems $P_1$ and $P_2$, the computational complexity consists of updating $\mathbf{U}$ and $\mathbf{V}$, where the computational complexity for $\mathbf{V}$ is $O(ncr)$ and the computational complexity for $\mathbf{U}$ is $O(nc^2r^2)$. For the problem $P_3$, the computation complexity $\mathbf{P}$ is $O(K(2r^2n + mrn + mr^2))$. For the problem $P_4$ for updating $\mathbf{H}$, the computational complexity is $O(Kmrn + r^2m + r^2n + 2Krn)$. For the problem $P_5$, the computational complexity for $\mathbf{w}$ is $O(K(mrn + m^2n))$. Considering the iterative strategy of the algorithm, set the number of iterations as $T$, the whole computational complexity is the $T$ times of the sum of computational complexity of the above five problems.

## IV. EXPERIMENTS

### A. Experimental Settings

Extensive experimental results are presented in this section to validate the performance of the proposed HSS-MVFC. The HSS-MVFC was compared with eight multi-view algorithms: MVKKM [32], MVSpec [32] and WV-CoFCM [6], Co-FCM [28], Co-FKM [27], TW-K-means [33], JNMF [36], MVKSC [41] and two single-view clustering algorithms: K-means and FCM. For single-view clustering algorithms, we constructed a single-view dataset by combining features from all visible views.

Two evaluation indices are used in this paper to evaluate the proposed algorithm.

1) Normalized Mutual Information (NMI) [8, 34]:

$$\text{NMI} = \frac{\sum_{i=1}^{c}\sum_{j=1}^{c} n_{i,j} \log n \cdot n_{i,j} / n_i \cdot n_j}{\sqrt{\sum_{i=1}^{c} n_i \log n_i / n \cdot \sum_{j=1}^{c} n_j \log n_j / n}} \quad (21)$$

where $n_{i,j}$ represents the number of samples matching the ith cluster with the real $j$ class, $n_i$ represents the number of samples contained in the ith cluster, $n_j$ represents the number of samples in the real jth class and $n$ represents the number of samples in the whole dataset.

2) Rand Index (RI) [8]:

$$\text{RI} = \frac{f00 + f11}{N(N-1)/2} \quad (22)$$

where $f00$ is the number of samples that has different real class labels and belongs to different clusters. $f11$ is the number of samples that has the same class labels and belong to the same cluster. $N$ is the number of samples.

The range of the above two evaluation indices is [0, 1], and the closer to 1, the better the performance are. The reported evaluation indices are the mean and variance of the results obtained from running 10 times under specific parameter settings. Table II shows the parameters of the adopted algorithms and the corresponding grid search range. For the algorithm JNMF, the parameters are set as the optimal values given in the original paper [36].

### B. Description of Multi-view Datasets

In the present study, six real world multi-view datasets are used to test the proposed algorithm. The detail of these datasets are shown in Table III.

### C. Performance Analysis

Based on the above datasets, the clustering results of eight algorithms are shown in Table IV and Table V. From Table IV and Table V we have the following observation: 1) The RI indices and NMI indices of HSS-MVFC are the highest in most dataset. 2) The mean values of HSS-MVFC are highest in all datasets. Thus, compared with classical methods, the proposed HSS-MVFC has certain advantages.

TABLE II
PARAMETERS OF THE ADOPTED ALGORITHMS AND THE CORRESPONDING GRID SEARCH RANGE

| Algorithms | Parameters and grid search range |
|---|---|
| MVKKM and MVSpec [20] | Index $p$: {1,1.1,…,1.9,2,3,…,6,7}. |
| Co-FKM [19] | Fuzzy index $m$: $m = \min(n, d-1)/(\min(n, d-1) - 2)$, where $d$ ($d > 3$) and $n(n) > 3$ are the dimensions and sample numbers, respectively. When $d \leq 3$, $m \in \{1,1.1,…,1.9,2,3,…,6,7\}$. |
| Co-FCM [17] | Fuzzy index $m$: same to Co-FKM. |
| TW-K-means [23] | Regularization parameter $\lambda$: {1,2,3,...,30}.<br>Regularization parameter $\eta$: {10,20,30,...,120}. |
| MV-Co-FCM [6] | Fuzzy index $m$: same to Co-FKM<br>Regularization parameter $\lambda$: {1e$^{-7}$,1e$^{-6}$,1e$^{-5}$,1e$^{-4}$,1e$^{-3}$,1e$^{-2}$,1e$^{-1}$,1e$^{0}$,1e$^{1}$,1e$^{2}$,1e$^{3}$,1e$^{4}$,1e$^{5}$,1e$^{6}$,1e$^{7}$}. |
| MVKSC [41] | Kernel parameter: {$2^{-6},2^{-5},1e^{-4},2^{-3},2^{-2},2^{-1},2^{0},2^{1},2^{2},2^{3},2^{4},2^{5},2^{6}$}.<br>Regularization parameter: {$2^{-6},2^{-5},1e^{-4},2^{-3},2^{-2},2^{-1},2^{0},2^{1},2^{2},2^{3},2^{4},2^{5},2^{6}$}. |
| HSS-MVFC | Fuzzy index $m$: same to Co-FKM<br>Regularization parameter $\lambda$: $\lambda \in \{2^{-3},2^{-2},2^{-1},2^{0},2^{1},2^{2},2^{3},2^{4},2^{5},2^{6},2^{7},2^{8},2^{9},2^{10},2^{11},2^{12},2^{13},2^{14}\}$<br>Regularization parameter $\eta$: $\eta \in \{1e^{-7},1e^{-6},1e^{-5},1e^{-4},1e^{-3},1e^{-2},1e^{-1},1e^{0},1e^{1},1e^{2},1e^{3},1e^{4},1e^{5},1e^{6},1e^{7}\}$<br>Low rank parameters $r$: $r \in \{10,20,30,…,100\}$, if $d_{\min}$ lower than 100, then $r \in \{10,20,30,…,d_{\min}\}$. |

TABLE III
DESCRIPTION OF MULTI-VIEW DATASETS

| Datasets | Samples | Features | Clusters | Description | |
|---|---|---|---|---|---|
| WebKB | 226 | 3104=2500+215+389 | 4 | WebKB is composed of web pages collected from computer science department of Washington university | View1: the text on web pages<br>View2: the anchor text in hyperlinks<br>View3: the text in its title |
| MF | 2000 | 123=76+47 | 10 | Handwritten digits represented by multiple features | View 1: Fourier coefficients<br>View 2: Zernike moments |
| Cora | 2708 | 4140=1433+2708 | 7 | Scientific publication classification | View1: Content<br>View2: Cites |
| Reuters | 1200 | 4000=2000+2000 | 6 | Document classification | View1: English<br>View2: French |
| WTP | 527 | 38=22+16 | 13 | daily measures of sensors in an urban waste water treatment plant | View1: Features of input conditions<br>View2: Features of output conditions |
| Dematology | 366 | 34=12+22 | 6 | Eryhemato-Squamous diseases in dermatology | View1: Clinical attributes<br>View2: Histopathological attributes |

TABLE IV
THE RI INDEX OF EACH ALGORITHM

| Datasets | K-means | FCM | MVKKM | MVSpec | Co-FKM | Co-FCM | TW-K-means | MV-Co-FCM | JNMF | MVKSC | HSS-MVFC |
|---|---|---|---|---|---|---|---|---|---|---|---|
| WebKB | 0.4606 (0.1144) | 0.7068 (0) | 0.7895 (0.0172) | 0.7080 (0.0102) | 0.8177 (0.0001) | 0.7569 (0.0227) | 0.7890 (0.0056) | 0.7954 (0.0091) | 0.6370 (0.0002) | 0.8196 (0.0000) | 0.8261 (0.0040) |
| MF | 0.9135 (0.0096) | 0.9124 (0.0076) | 0.9246 (0.0001) | 0.9246 (0.0031) | 0.9005 (0.0022) | 0.9202 (0.0006) | 0.9235 (0.0070) | 0.9218 (0.0015) | 0.9158 (0.0028) | 0.8762 (0.0083) | 0.9381 (0.0016) |
| Cora | 0.4641 (0.2567) | 0.6670 (0.0527) | 0.7519 (0.0051) | 0.4763 (0.0501) | 0.7437 (0.0059) | 0.7328 (0.0071) | 0.5894 (0.0329) | 0.7240 (0.0080) | 0.7396 (0.0146) | 0.7709 (0.0000) | 0.7615 (0.0095) |
| Reuters | 0.5003 (0.2014) | 0.5359 (0.0243) | 0.6520 (0.0173) | **0.7776 (0.0002)** | 0.7286 (0.0115) | 0.7237 (0.0129) | 0.7399 (0.0049) | 0.7412 (0.0063) | 0.7612 (0.0044) | 0.7450 (0.0013) | 0.7725 (0.0021) |
| WTP | 0.7026 (0.0054) | 0.7063 (0.0015) | 0.6990 (0.0012) | 0.7015 (0.0001) | 0.7080 (0.0087) | 0.7070 (0.0021) | 0.7093 (0.0044) | 0.7074 (0.0041) | 0.7061 (0.0056) | 0.6985 (0.0000) | 0.7165 (0.0002) |
| Dermatology | 0.8389 (0.0605) | 0.9037 (0.0286) | 0.9169 (0.0001) | 0.8993 (0.0001) | 0.9536 (0.0041) | 0.9556 (0.0006) | 0.9033 (0.0407) | 0.9558 (0.0041) | 0.8785 (0.0013) | 0.7041 (0.0140) | 0.9783 (0.0001) |
| Mean | 0.6467 | 0.7387 | 0.7890 | 0.7479 | 0.8087 | 0.7994 | 0.7757 | 0.8076 | 0.7730 | 0.7691 | 0.8322 |

TABLE V
THE NMI INDEX OF EACH ALGORITHM

| Datasets | K-means | FCM | MVKKM | MVSpec | Co-FKM | Co-FCM | TW-K-means | MV-Co-FCM | JNMF | MVKSC | HSS-MVFC |
|---|---|---|---|---|---|---|---|---|---|---|---|
| WebKb | 0.1663 (0.2003) | 0.4648 (0.0009) | 0.5670 (0.0219) | 0.4539 (0.0132) | 0.5915 (0.0120) | 0.4785 (0.0282) | 0.5689 (0.0178) | 0.5519 (0.0156) | 0.4200 (0.0001) | 0.5825 (0.0000) | 0.6385 (0.0279) |
| MF | 0.6386 (0.0314) | 0.6256 (0.0230) | 0.6668 (0.0021) | 0.6948 (0.0028) | 0.5254 (0.0133) | 0.6378 (0.0024) | 0.6883 (0.0202) | 0.6724 (0.0099) | 0.6254 (0.0165) | 0.4323 (0.0206) | 0.7426 (0.0124) |
| Cora | 0.1056 (0.0755) | 0.0961 (0.0117) | 0.1961 (0.0314) | 0.2000 (0.0106) | 0.1512 (0.0075) | 0.1322 (0.0127) | 0.0969 (0.0317) | 0.2380 (0.0141) | 0.2976 (0.0174) | 0.2770 (0.0035) | 0.2903 (0.0099) |
| Retuters | 0.1888 (0.0989) | 0.0978 (0.0159) | 0.2836 (0.0012) | 0.3127 (0.0015) | 0.2585 (0.0127) | 0.2360 (0.0186) | 0.2579 (0.0078) | 0.2680 (0.0141) | 0.2691 (0.0075) | 0.1941 (0.0172) | 0.3523 (0.0083) |
| WTP | 0.1943 (0.0224) | 0.1967 (0.0083) | 0.1102 (0.0123) | 0.1478 (0.0001) | 0.1980 (0.0087) | 0.1950 (0.0015) | 0.2101 (0.0165) | 0.1964 (0.0122) | 0.1809 (0.0200) | 0.1163 (0.0000) | 0.2369 (0.0040) |
| Dermatology | 0.7709 (0.0769) | 0.8619 (0.0441) | 0.8837 (0.0001) | 0.7721 (0.0001) | 0.8522 (0.0041) | 0.8705 (0.0004) | 0.8428 (0.0479) | 0.8799 (0.0020) | 0.7878 (0.0043) | 0.8910 (0.0010) | 0.9244 (0.0001) |
| Mean | 0.3441 | 0.3905 | 0.4512 | 0.4302 | 0.4295 | 0.4250 | 0.4441 | 0.4678 | 0.4301 | 0.4155 | 0.5308 |

## D. Algorithm Convergence

Fig. 3 shows the convergence curves of the proposed HSS-MVFC in each dataset. It is observed that when the number of iteration is less than 200, the loss function fell sharply, and when the number of iteration is close to 800, the algorithm converges.

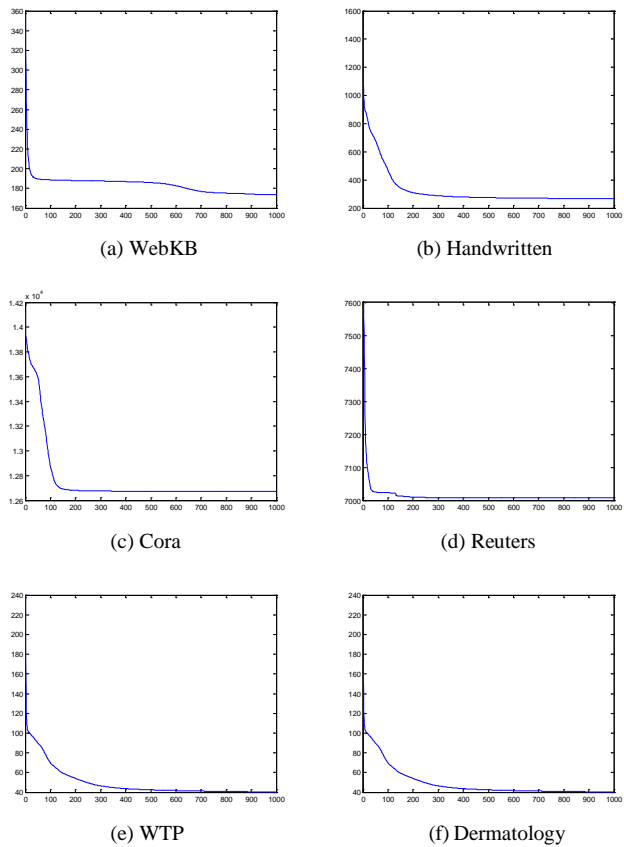

(a) WebKB  (b) Handwritten
(c) Cora  (d) Reuters
(e) WTP  (f) Dermatology

Fig.4. Convergence curves of HSS-MVFC

## E. Statistical Analysis

The main statistical analysis that we performed was Friedman test combined with post-hoc Holm test, where Friedman test [35] was used to check if the difference among the 11 algorithms is statistically significant. The null hypothesis, which says that the clustering performances of all methods were the same. The rejection of the null hypothesis means that the clustering performances of the 11 algorithms had statistically significant difference. The significance level ($\alpha$) is set as 0.05. Table VI and Table VII show the rankings of the 11 algorithms based on RI and NMI, respectively, in which a lower ranking indicates a better performance. Clearly, HSS-MVFC had the best performance. And the p-values of them are both lower than 0.05, which means that the 11 algorithms had statistically significant difference *w.r.t.* RI and NMI.

The Holm test [42] was then used to compare the best method, i.e., HSS-MVFC with the other 10 algorithms. The null hypothesis, i.e., there does not exist statistically significant difference between two algorithms, is rejected if the *p*-value is below Holm value. The Holm value is calculated with $\alpha(k-1)$ and $\alpha$ is set as 0.05. For the index RI, Table VIII shows that there was statistically significant difference between HSS-MVFC and K-means, FCM, JNMF, MVKSC, MVSpec. For the index NMI, Table IX shows that there was statistically significant difference between HSS-MVFC and K-means, FCM, Co-FCM, JNMF, MVKSC, MVSpec. Although the null hypothesis was not rejected for some algorithms as shown in Table VIII and Table IX, Tables VI and Table VII show that HSS-MVFC still outperformed them slightly.

## V. CONCLUSIONS

Aiming at multi-view data clustering analysis, a multi-view fuzzy clustering method based on shared hidden space learning is proposed. By introducing the non-negative matrix factorization, the characteristic matrix in each view space is decomposed into the base matrix and the coefficient matrix, so that the multi-view data can be considered as the projection of the common data in the hidden space of each view. A new multi-view sharing hidden space learning strategy is used to find the correlation between different views. In the multi-view clustering process, the information of the visible view is considered in order to achieve the purpose of discovering the essential attribute of the clustering object, and thus obtain more reasonable clustering results based on these essential attributes. In addition, the multi-view adaptive weighting strategy based on Shannon entropy also effectively realizes the adaptive co-ordination of different views. Our experimental results on multiple benchmark datasets show that the proposed method has better clustering performance than previous multi-view algorithms and related single view algorithms.

TABLE VI
FRIEDMAN TEST BASED ON RI INDEX

| Algorithm | Ranking | p-value | Hypothesis |
|---|---|---|---|
| K-means | 9.8333 | | |
| FCM | 8 | | |
| MVKKM | 5.75 | | |
| MVSpec | 6.4167 | | |
| Co-FKM | 5.1667 | | |
| Co-FCM | 5.8333 | 0.006361 | Reject |
| TW-K-means | 5.6667 | | |
| MV-Co-FCM | 4.5 | | |
| JNMF | 6.8333 | | |
| MVKSC | 6.6667 | | |
| HSS-MVFC | 1.3333 | | |

TABLE VII
FRIEDMAN TEST BASED ON NMI INDEX

| Algorithm | Ranking | p-value | Hypothesis |
|---|---|---|---|
| K-means | 9 | | |
| FCM | 8 | | |
| MVKKM | 5.5 | | |
| MVSpec | 61667 | | |
| Co-FKM | 5.8333 | 0.015895 | Reject |
| Co-FCM | 6.8333 | | |
| TW-K-means | 5.6667 | | |
| MV-Co-FCM | 4.6667 | | |
| JNMF | 6.8333 | | |
| MVKSC | 6.3333 | | |
| HSS-MVFC | 1.1667 | | |

TABLE VIII
POST-HOC TEST BASED ON RI INDEX

| $i$ | Algorithms | $z=(R_0-R_i)/SE$ | p-value | Holm= $\alpha/i$ | Hypothesis |
|---|---|---|---|---|---|
| 10 | K-means | 4.43898 | 0.000009 | 0.005 | Reject |
| 9 | FCM | 3.481553 | 0.000499 | 0.005556 | Reject |
| 8 | JNMF | 2.872281 | 0.004075 | 0.00625 | Reject |
| 7 | MVKSC | 2.785242 | 0.005349 | 0.007143 | Reject |
| 6 | MVSpec | 2.654684 | 0.007938 | 0.008333 | Reject |
| 5 | Co-FCM | 2.350048 | 0.018771 | 0.01 | Not Reject |
| 4 | MVKKM | 2.306529 | 0.021081 | 0.0125 | Not Reject |
| 3 | TW-K-means | 2.26301 | 0.023635 | 0.016667 | Not Reject |
| 2 | Co-FKM | 2.001893 | 0.045296 | 0.025 | Not Reject |
| 1 | MV-Co-FCM | 1.653738 | 0.098181 | 0.05 | Not Reject |

TABLE IX
POST-HOC TEST BASED ON NMI INDEX

| $i$ | Algorithms | $z=(R_0-R_i)/SE$ | p-value | Holm= $\alpha/i$ | Hypothesis |
|---|---|---|---|---|---|
| 8 | K-means | 4.090825 | 0.000043 | 0.005 | Reject |
| 7 | FCM | 3.568592 | 0.000359 | 0.005556 | Reject |
| 6 | Co-FCM | 2.95932 | 0.003083 | 0.00625 | Reject |
| | JNMF | 2.95932 | 0.003083 | 0.007143 | Reject |
| | MVKSC | 2.698204 | 0.006971 | 0.008333 | Reject |
| 5 | MVSpec | 2.611165 | 0.009023 | 0.01 | Reject |
| 4 | Co-FKM | 2.437087 | 0.014806 | 0.0125 | Not Reject |
| 3 | TW-K-means | 2.350048 | 0.018771 | 0.016667 | Not Reject |
| 2 | MVKKM | 2.26301 | 0.023635 | 0.025 | Not Reject |
| 1 | MV-Co-FCM | 1.827815 | 0.067577 | 0.05 | Not Reject |

## VI. APPENDIX

The more detailed derivation process for the optimization method is demonstrated as follows. For (16), the derivation of $P_2(\mathbf{V})$ w.r.t. $\mathbf{V}$ is as follows:

$$\frac{J(\mathbf{U},\mathbf{V},\mathbf{P}^k,\mathbf{H},\mathbf{w})}{\partial v_l} = \sum_{i=1}^{n} u_{li}^m h_i - \sum_{i}^{n} u_{li}^m v_l \quad (A1)$$

$$\mathbf{v}_l = \sum_{i=1}^{n} u_{li}^m \mathbf{h}_i \bigg/ \sum_{i=1}^{n} u_{li}^m \quad (A2)$$

For (17), the derivation of $P_1(\mathbf{U})$ w.r.t. $\mathbf{U}$ is as follows:

$$\frac{J(\mathbf{U},\mathbf{V},\mathbf{P}^k,\mathbf{H},\mathbf{w})}{\partial u_{li}} = m u_{li}^{m-1} \|h_i - v_l\| + \mu_i = 0 \quad (A3)$$

Due to $\sum_{l=1}^{c} u_{li} = 1$, $1 \leq i \leq n$, we can conclude that.

$$\mu_i = -\sum_{l'=1}^{c} (m \|h_i - v_{l'}\|^2) \quad (A4)$$

$$u_{li} = 1 \bigg/ \sum_{l'=1}^{c} (\frac{\|\mathbf{h}_i - \mathbf{v}_l\|^2}{\|\mathbf{h}_i - \mathbf{v}_{l'}\|^2})^{\frac{2}{m-1}}, \quad 1 \leq i \leq n \quad (A5)$$

For $k$th views, the derivation of $P_3(\mathbf{P}^k)$ w.r.t. $\mathbf{P}^k$ is as follows:

$$\frac{J(\mathbf{U},\mathbf{V},\mathbf{P}^k,\mathbf{H},\mathbf{w})}{\partial \mathbf{P}^k} = 2\mathbf{P}^k(H^T H) - 2X^k H^T \quad (A5)$$

$$(\mathbf{P}^k)_{i,j}^{(t+1)} \leftarrow \left[\frac{(\mathbf{X}^k \mathbf{H}^T)_{i,j}}{((\mathbf{P}^k)^{(t)} \mathbf{H}\mathbf{H}^T)_{i,j}}\right](\mathbf{P}^k)_{i,j}^{(t)} \quad (A6)$$

For (19), the gradient descent method is used to optimize $\mathbf{H}$, then the iterative formula is as follows:

$$(\mathbf{H})_{i,j}^{(t+1)} = \mathbf{H}_{i,j}^{(t)} - step \cdot \frac{\partial P_4(\mathbf{H})}{\partial (\mathbf{H})_{i,j}} \quad 1 \leq j \leq r, \ 1 \leq i \leq n \quad (A7)$$

where

$$\frac{\partial P(\mathbf{H})}{\partial (\mathbf{H})_{i,j}} = 2\sum_{l=1}^{C} u_{li}^{m}((\mathbf{H})_{i,j} - (\mathbf{V})_{lj}) +$$
$$\lambda \sum_{k=1}^{K}(2w_k \sum_{h=1}^{m_k}(-(\mathbf{P})_{h,j}^{k}(\mathbf{X})_{h,i}^{k} + (\mathbf{P}^k \mathbf{H}^{(t)})_{h,i}(\mathbf{P})_{h,j}^{k})) \quad \text{(A8)}$$
$$= 2\sum_{l=1}^{C} u_{li}^{m}(\mathbf{H})_{i,j} - 2\sum_{l=1}^{C} u_{li}^{m}(\mathbf{V})_{lj} +$$
$$2\lambda \sum_{k=1}^{K} w_k (-(\mathbf{P}^k)^T \mathbf{X}^k) + 2\lambda \sum_{k=1}^{K} w_k (\mathbf{P}^k)^T \mathbf{P}^k \mathbf{H}^{(t)}$$

and the step is set as follows:

$$step = \frac{(\mathbf{H})_{i,j}^{(t)}}{2\sum_{l=1}^{c} u_{li}^{m}(\mathbf{H})_{i,j} + 2\lambda \sum_{k=1}^{K} w_k (\mathbf{P}^k)^T \mathbf{P}^k \mathbf{H}^{(t)}} \quad \text{(A9)}$$

By substituting (A8) and (A9) into (A7), the update rule for $\mathbf{H}$ is as follows:

$$(\mathbf{H})_{i,j}^{(t+1)} = \frac{\sum_{l=1}^{C} u_{li}^{m}(\mathbf{V})_{lj} + \lambda \sum_{k=1}^{K} w_k (\mathbf{P}^k \mathbf{X}^{(t)})_{i,j}}{\sum_{l=1}^{C} u_{li}^{m}(\mathbf{H})_{i,j}^{(t)} + \lambda \sum_{k=1}^{K} w_k (\mathbf{P}^k (\mathbf{P}^k)^T \mathbf{H}^{(t)})_{i,j}} (\mathbf{H})_{i,j}^{(t)} \quad \text{(A10)}$$
$$1 \le j \le r, \ 1 \le i \le n$$

REFERENCE

[1] S. Bickel, T. Scheffer, "Multi-view clustering". in Proc. *IEEE International Conference on Data Mining*, pp. 19-26, 2004

[2] X. Cai, F. Nie, H. Huang, "Multi-view K-means clustering on big data". in Proc. *International Joint Conference on Artificial Intelligence* pp. 2598-2604, 2013

[3] A. Kumar, P. Rai, "Co-regularized multi-view spectral clustering". in *Proc. International Conference on Neural Information Processing Systems*, pp. 1413-1421, 2011

[4] A. Kumar, H.D. Iii, "A co-training approach for multi-view spectral clustering". in *Proc. International Conference on International Conference on Machine Learning* pp. 393-400, 2011

[5] D. Zhou, C.J.C. Burges, "Spectral clustering and transductive learning with multiple views". in *Proc. Machine Learning, Proceedings of the Twenty-Fourth International Conference* pp. 1159-1166, 2007

[6] Y. Jiang, F.L. Chung, S. Wang, et al. "Collaborative fuzzy clustering from multiple weighted views". *IEEE Transactions on Cybernation*, vol. 45, no. 3, pp. 688-701, 2015

[7] S. Yu, L.C. Tranchevent, X. Liu, et al. "Optimized data fusion for kernel k-means clustering". *IEEE Transactions on Pattern Analysis & Machine Intelligence*, vol. 34, no. 5, pp. 1031-1039, 2012

[8] L. Jing, M.K. Ng, J.Z. Huang, "An Entropy Weighting k-Means Algorithm for Subspace Clustering of High-Dimensional Sparse Data". *IEEE Transactions on Knowledge & Data Engineering*, vol. 19, no. 8, pp. 1026-1041, 2007

[9] L. Zhu, F.L. Chung, S. Wang, "Generalized Fuzzy C-Means Clustering Algorithm With Improved Fuzzy Partitions". *IEEE Transactions on Systems Man & Cybernetics Part B Cybernetics A Publication of the IEEE Systems Man & Cybernetics Society*, vol. 39, no. 3, pp. 578-591, 2009

[10] L.O. Hall, D.B. Goldgof, "Convergence of the Single-Pass and Online Fuzzy C-Means Algorithms". *IEEE Transactions on Fuzzy Systems*, vol. 19, no. 4, pp. 792-794, 2011

[11] Z. Deng, S. Wang, X. Wu, et al. "Robust Maximum Entropy Clustering Algorithm RMEC and Its Outlier Labeling". *Engineering Science*, vol. 6, no. 9, pp. 38-45, 2004

[12] N.B. Karayiannis, "MECA: maximum entropy clustering algorithm". in *Proc. IEEE World Congress on Computational Intelligence*, vol.631, pp. 630-635, 1994

[13] R. Krishnapuram, J.M. Keller, "A possibilistic approach to clustering". *IEEE Transactions on Fuzzy Systems*, vol. 1, no. 2, pp. 98-110, 2002

[14] R. Krishnapuram, J.M. Keller, "The possibilistic C-means algorithm: insights and recommendations". *IEEE Transactions on Fuzzy Systems*, vol.4, no.3, pp. 385-393, 2002

[15] S. Asur, D. Ucar, S. Parthasarathy, "An ensemble framework for clustering protein–protein interaction networks". *Bioinformatics*, vol. 23, no. 13, pp. i29, 2007

[16] H. Wang, H. Shan, A. Banerjee, "Bayesian cluster ensembles". *Statistical Analysis & Data Mining the Asa Data Science Journal*, vol. 4, no. 1, pp. 54-70, 2011

[17] M.B. Blaschko, C.H. Lampert, "Correlational spectral clustering". in *Proc. Computer Vision and Pattern Recognition*, pp. 1-8, 2008

[18] E. Bruno, S. Marchand-Maillet, "Multiview clustering: a late fusion approach using latent models". in *Proc. International ACM SIGIR Conference on Research and Development in Information Retrieval* pp. 736-737, 2009

[19] X. Liu, S. Ji, W. Glänzel, et al. "Multiview Partitioning via Tensor Methods". *IEEE Transactions on Knowledge & Data Engineering*, vol. 25, no. 5, pp. 1056-1069, 2013

[20] C. Zhang, H. Fu, S. Liu, et al. "Low-Rank Tensor Constrained Multiview Subspace Clustering". in *Proc. IEEE International Conference on Computer Vision* pp. 1582-1590, 2015

[21] Y. Li, F. Nie, H. Huang, et al. "Large-scale multi-view spectral clustering via bipartite graph". in *Proc. Twenty-Ninth AAAI Conference on Artificial Intelligence* pp. 2750-2756, 2015

[22] S.Y. Li, Y. Jiang, Z.H. Zhou, "Partial multi-view clustering". in *Proc. AAAI Conference on Artificial Intelligence*, 2010

[23] H. Wang, F. Nie, H. Huang, "Multi-view clustering and feature learning via structured sparsity". in *Proc. International Conference on Machine Learning* pp. 352-360, 2013

[24] X. Xie, S. Sun, "Multi-view clustering ensembles". in *Proc. International Conference on Machine Learning and Cybernetics,* pp. 51-56, 2014

[25] P. Summers, G. Menegaz, "Multiview cluster ensembles for multimodal MRI segmentation" *International Journal of Imaging Systems & Technology*, vol.25, no. 1, pp.56-67, 2015

[26] S.F. Hussain, M. Mushtaq, Z. Halim, "Multi-view document clustering via ensemble method" *Journal of Intelligent Information Systems*, vol.43, no. 1, pp 81-99, 2014

[27] G. Cleuziou, M. Exbrayat, L. Martin, et al. "CoFKM: A Centralized Method for Multiple-View Clustering". in *Proc. Ninth IEEE International Conference on Data Mining* pp. 752-757, 2009

[28] W. Pedrycz, "Collaborative fuzzy clustering". *Pattern Recognition Letters*, vol.23, no. 14, pp. 1675-1686, 2002

[29] S.M. Kakade, D.P. Foster, "Multi-view Regression Via Canonical Correlation Analysis". *Lecture Notes in Computer Science*, 4539, pp. 82-96, 2007

[30] Li, T.,Ding, C. Non-negative matrix factorization for clustering: A survey, 2013

[31] X. Zhang, L. Zhao, L. Zong, et al. "Multi-view Clustering via Multi-manifold Regularized Nonnegative Matrix Factorization". *Neural networks : the official journal of the International Neural Network Society*, 88: pp. 74, 2017

[32] G. Tzortzis, A. Likas, "Kernel-Based Weighted Multi-view Clustering". in *Proc. IEEE International Conference on Data Mining,* pp. 675-684, 2012

[33] X. Chen, X. Xu, J.Z. Huang, et al. "TW-k-means: Automated two-level variable weighting clustering algorithm for multiview data". *IEEE Transactions on Knowledge & Data Engineering*, vol. 25, no. 4, pp. 932-944, 2013

[34] Z. Deng, K.S. Choi, F.L. Chung, et al. "Enhanced soft subspace clustering integrating within-cluster and between-cluster information". *Pattern Recognition*, vol. 43, no.3, pp. 767-781, 2010

[35] J. Demšar, "Statistical comparisons of classifiers over multiple data sets". *Journal of Machine Learning Research*, vol.7, no.1, pp. 1-30, 2006

[36] Gao J, Han J, Liu J, et al. Multi-view clustering via joint nonnegative matrix factorization[C]. siam international conference on data mining, 2013: 252-260.

[37] Shao W , He L , Yu P S . Multiple Incomplete Views Clustering via Weighted Nonnegative Matrix Factorization with L2,1 Regularization[C]// ECML/PKDD. Springer, 2015.

[38] Zong L , Zhang X , Zhao L , et al. Multi-view clustering via multi-manifold regularized non-negative matrix factorization[J]. Neural Networks, 2017, 88:74-89.

[39] Zhang, X., Gao, H., Li, G., Zhao, J., Huo, J., Yin, J., … Zheng, L. (2018). Multi-view clustering based on graph-regularized nonnegative matrix factorization for object recognition. Information Sciences, 432, 463–478.

[40] Tan Y, Ou W, Long F, et al. Multi-view Clustering via Co-regularized Nonnegative Matrix Factorization with Correlation Constraint[C] International Conference on Cloud Computing & Big Data. IEEE, 2017.

[41] Houthuys L, Langone R, Suykens J A K. Multi-View Kernel Spectral Clustering[J]. Information Fusion, 2018, 44:46-56

[42] S. García, A. Fernández, Luengo J, et al. A study of statistical techniques and performance measures for genetics-based machine learning: accuracy and interpretability[J]. Soft Computing, 2009, 13(10):959-977.